\pdfoutput=1
\documentclass[10pt,twocolumn,a4paper]{article}
\usepackage[margin=1.8cm,columnsep=0.7cm]{geometry}
\usepackage{amsmath,amssymb}
\usepackage{graphicx,booktabs,tabularx,array}
\usepackage{dblfloatfix}

\usepackage[hidelinks]{hyperref}
\usepackage{xurl}
\usepackage[font=small,labelfont=bf]{caption}
\usepackage{titlesec}
\titleformat{\section}{\large\bfseries}{\thesection}{0.6em}{}
\titleformat{\subsection}{\normalsize\bfseries}{\thesubsection}{0.6em}{}
\AtBeginDocument{\setlength{\abovedisplayskip}{6pt plus 2pt}\setlength{\belowdisplayskip}{6pt plus 2pt}}

\title{\Large\bfseries The Hidden Look-Ahead:\\ Point-in-Time Leakage in Electricity Market Data}
\author{Abdelkrim Alfalah\\[2pt]Goldsmiths, University of London\\[2pt]\small \href{mailto:a.alfalah@gold.ac.uk}{a.alfalah@gold.ac.uk}\\[2pt]\small ORCID:~\href{https://orcid.org/0009-0008-2889-5033}{0009-0008-2889-5033}}
\date{}
\begin{document}
\twocolumn[
\begin{@twocolumnfalse}
\maketitle
\begin{abstract}
Empirical studies of electricity prices routinely state that their models use only information available at the time of the trading decision. For weather-driven forecasts drawn from commercial archives, this condition is harder to meet than it appears, and whether it holds can change over time. We identify three timing pitfalls. A forecast run is labelled by the issue time of its weather model, but becomes available hours later and keeps changing after publication. Auction gates follow local clock time, while archives use UTC. Outturn series are revised after publication without earlier values being kept. Using the timestamps of 128,656 forecast runs from ten weather-model configurations, matched to the day-ahead and first intraday auction gates in Great Britain over 2022--2026, we find that selecting ``the latest run issued before the gate'' picks a run not yet published at the gate on more than 99\% of delivery days for six of the ten configurations at one of the two gates. For GFS at the day-ahead gate, the share is 0.6\% in winter and 99.8\% in summer. The exposure also shifts as publication delays drift, and an action margin of a few minutes can cost a full six-hour forecast cycle. The wrongly selected inputs differ materially from those available at the gate, but their apparent accuracy gain ranges from $-$9\% to +24\% across variables, models, gates and seasons, so its effect cannot be assumed and must be measured. We argue that point-in-time correctness is a dynamic property of the data pipeline, and propose a protocol that audits it from timestamps before any model is fitted.
\end{abstract}
\vspace{1em}
\end{@twocolumnfalse}
]
\section{Introduction}

Electricity in Great Britain, as in most European markets, is traded for physical delivery through a sequence of auctions held on the day before delivery and on the delivery day itself, followed by continuous intraday trading. With a growing share of wind and solar generation, the information that moves prices between these auctions is increasingly weather-driven: forecasts of wind output, solar output and demand, updated several times a day as new numerical weather prediction runs become available (Weron, 2014; Lago et al., 2021). Empirical studies of day-ahead and intraday prices therefore rely heavily on such forecasts as explanatory variables.

A study of this kind is only informative about trading decisions if each feature was available when the corresponding decision had to be taken. This point-in-time condition is easy to state and, as we show, easy to violate. Forecast archives make it so: runs are indexed by the time their weather model was initialised, and the obvious query, ``the latest run issued before the gate closure'', returns runs that were published only hours later. Sequential auctions make the problem sharper, because each gate closure defines a different information set, a few hours apart.

The problem concerns reconstructed history, not live trading. A system operating at the gate can only see runs that have been published, so it cannot look ahead. Leakage arises when the past is rebuilt from a complete archive, as in backtests, model training and empirical research, which then receive fresher or later-revised inputs than any participant had. Its most direct consequence is a training--serving skew: a model trained on runs that were not yet available at the gate receives older runs in production, and can perform worse for no visible reason.

This paper examines how point-in-time forecast datasets are constructed, and how often a natural construction fails. We match the publication timestamps of 128,656 forecast runs to the gate closures of the Great Britain day-ahead and intraday auctions over 2022--2026, and measure how often each of three selection rules uses information that was not yet available.

The paper makes four contributions. First, we identify and formalise three timing pitfalls: confusing a run's issue time with its publication time; stating gate closures in UTC when the auctions follow local clock time; and using outturn series that are revised after publication without the earlier values being retained. We distinguish two mechanisms: the use of runs that had not been published, and exposure to values revised after the gate. Second, we measure the resulting exposure on a large commercial archive. For six of the ten configurations, selecting the latest run issued before the gate picks an unpublished run on more than 99\% of delivery days at one of the two gates. The exposure depends on the weather-model configuration, the gate and the season, not on the variable forecast, and it shifts as publication delays drift and as the action margin crosses thresholds as short as 5 minutes. Third, we show that the wrongly selected inputs differ materially from those available at the gate, but that their effect on apparent forecast accuracy ranges from negative to strongly positive, so it cannot be assumed. Fourth, we propose a protocol, validated by direct observation of publication times, that audits a dataset from timestamps before any model is fitted. Together, the results support one thesis: point-in-time correctness is a dynamic property of the data pipeline, not a one-off check.

The remainder of the paper is organised as follows. Section 2 reviews related work. Section 3 describes the auction setting and defines the life cycle of a forecast run and the three selection rules. Section 4 describes the data. Section 5 reports the results for forecast runs, including how much the wrongly selected inputs differ from those available at the gate, and Section 6 extends the analysis to prices and outturns. Section 7 sets out the recommended protocol, Section 8 discusses limitations and future work, and Section 9 concludes.

\section{Related work}

\textbf{Real-time data and vintages.} Macroeconomics has long distinguished between data as first published and data as later revised. Croushore and Stark (2001) built a real-time data set of quarterly vintages, and Stark and Croushore (2002) showed that forecast errors can be deceptively low when models are evaluated on revised rather than real-time data. Croushore (2011) reviews the literature that followed. Our treatment of outturn series (Section 6) applies the same idea to electricity data. The forecast-run problem is an analogue with one difference: forecast archives retain every vintage, so the problem lies in selecting the right one rather than in reconstructing it.

\textbf{Temporal data management.} Database research separates the \emph{valid time} of a fact, when it holds, from its \emph{transaction time}, when it was recorded (Snodgrass, 1999). In these terms, the naive rule treats an issue-time label as if it were a transaction time. Point-in-time selection requires the transaction time: the publication or finalisation time of a run.

\textbf{Leakage in predictive modelling.} Kaufman et al. (2012) define leakage as the use of information about the target that would not legitimately be available at prediction time, and propose separating the learning and prediction stages to avoid it. Kapoor and Narayanan (2023) survey leakage across 17 scientific fields that use machine learning and propose a taxonomy of eight types. The pitfalls studied here are instances of temporal leakage. They are specific in that they arise from metadata conventions rather than from modelling choices, and can therefore be detected from timestamps alone.

\textbf{Electricity price forecasting.} The standard reviews of electricity price forecasting (Weron, 2014; Lago et al., 2021) establish best practice for model comparison, evaluation metrics and open benchmarks. In the benchmark of Lago et al. (2021), exogenous inputs are day-ahead load and generation forecasts published by system operators and exchanges, with one day-ahead forecast per delivery day. The time at which these inputs become available relative to the gate is not part of the protocol. Studies of intraday prices are more explicit about information sets. Narajewski and Ziel (2020), for example, fix the forecast time at 3 h 15 min before delivery and use only information available at that moment: imbalance volumes enter with their 15-minute publication lag, and imbalance prices are left out because they are published two months later.

\textbf{Weather forecasts as inputs.} In operational renewable-energy forecasting, the delay between the initialisation of a weather model and the availability of its output is well known; weather centres publish it as part of their dissemination schedules (e.g. ECMWF, 2026). Research benchmarks often abstract from it. The recent R$^2$Energy benchmark (Sheng et al., 2026), for instance, uses reanalysis data as a proxy for future weather forecasts, so that all models have identical access to them. That design is appropriate for comparing models, but it sets publication timing aside.

To our knowledge, no study has measured when weather-driven forecast runs become available relative to the gate closures of electricity auctions, or shown how the resulting leakage varies with the weather model, the gate, the season and the passage of time.

\section{Setting and definitions}

\subsection{Sequential auctions and their gate closures}

Great Britain trades electricity for physical delivery in a sequence of auctions that close on the day before delivery (D$-$1) and on the delivery day itself (D). We describe three of them and analyse the first two. The hourly day-ahead auction operated by Nord Pool (N2EX) closes at 09:50 on D$-$1 (Nord Pool, 2023) and publishes results by 10:00 (Nord Pool, 2020). The first half-hourly intraday auction operated by EPEX SPOT (IDA1) closes at 17:30 on D$-$1 and publishes results from 18:00 (EPEX SPOT, 2024). The second intraday auction (IDA2) closes at 08:00 on D; it is held on the delivery day itself and is left for future work. All three timetables are defined in UK clock time (Nord Pool, 2023; EPEX SPOT, 2024): Greenwich Mean Time (GMT) in winter and British Summer Time (BST, UTC+1) in summer. The access rules of the GB--Norway interconnector, in force since 2021, likewise set the day-ahead gate at 09:50 UK time (National Grid NSL, 2021). The day-ahead gate closure moved from 11:00 to 09:50 on 1 January 2021 (Nord Pool, 2020), when Great Britain left the European day-ahead market coupling; we found no later change, so the period studied here (2022--2026) falls under a single timetable regime.

Each gate closure defines a decision point. A participant who submits orders to an auction can only use information available before the order book closes, less whatever time is needed to process that information and submit orders. Because the auctions are sequential, the information set grows from one gate to the next, and a feature that is admissible at a later gate may be inadmissible at an earlier one. Figure 1 places the three gates on a common time axis, in UTC, for a winter and a summer day.

\begin{figure*}[tp]
\centering
\includegraphics[width=\textwidth]{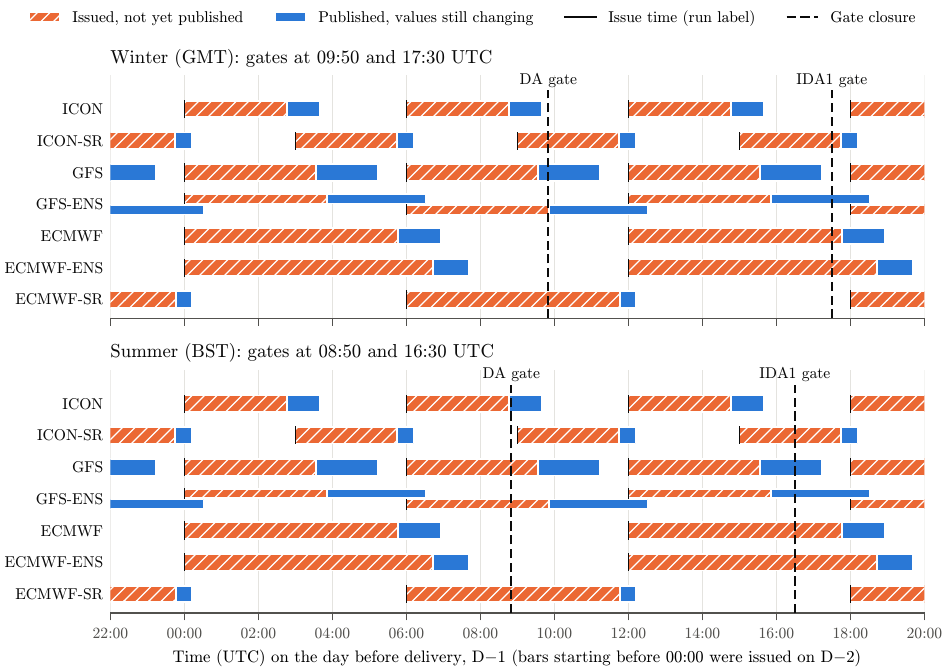}
\caption{Gate closures and the life cycle of forecast runs on the day before delivery, in UTC, for a winter day (top) and a summer day (bottom). Each bar is one forecast run. It starts at the run's issue time (black tick), remains unpublished until its publication time (orange, hatched), and is then available but still changing until its finalisation time (blue); after finalisation no bar is drawn. Bar lengths are the 2022--2026 median delays for each configuration. The dashed lines mark the N2EX day-ahead gate (09:50 UK clock time) and the EPEX IDA1 gate (17:30 UK clock time), which fall one hour earlier in UTC in summer. A gate line crossing an orange segment marks a run that the naive rule would select although it was not yet published; a gate line crossing a blue segment marks a run that was published but not yet final.}
\end{figure*}

\subsection{The life cycle of a forecast run}

Weather-driven forecasts of wind generation, solar generation and electricity demand are produced from numerical weather prediction (NWP) models. Each NWP model is initialised at fixed cycle times (typically 00, 06, 12 and 18 UTC; for some configurations at 03, 09, 15 and 21 UTC, or only twice a day). A forecast vendor converts each model run into power and demand forecasts and distributes the result as a \emph{forecast run}: a complete forecast trajectory identified by its weather model and its cycle time.

Three timestamps describe a forecast run $r$:

\begin{itemize}
\item the \textbf{issue time} $\tau^{I}_r$, the initialisation time of the underlying weather model run, which is also the label by which the run is identified and stored;
\item the \textbf{publication time} $\tau^{P}_r$, the moment the run first becomes available to clients;
\item the \textbf{finalisation time} $\tau^{F}_r$, the moment after which the values of the run no longer change.
\end{itemize}

By construction $\tau^{I}_r < \tau^{P}_r \le \tau^{F}_r$. The gap $\tau^{P}_r - \tau^{I}_r$ reflects the time needed to run the weather model, distribute its output and convert it into power forecasts; it is several hours (Section 5.1). The gap $\tau^{F}_r - \tau^{P}_r$ arises because a run is published progressively as later forecast steps are ingested, so its values continue to change after it first appears.

This study uses the forecast archive of a commercial energy-data vendor (Section 4).\footnote{Energy Quantified (www.energyquantified.com). Data retrieved through the vendor's Python client between June and October 2026.} The archive stores all three timestamps for every run, under different field names. The vendor's documentation defines them only briefly. We therefore validated the interpretation directly. We polled the vendor's interface every 60 seconds for one evening and recorded when each new run first became visible. Three runs (ICON, GFS and an ECMWF short-range run, all initialised at 18 UTC) appeared 1, 36 and 9 seconds after their recorded publication time, and never before it. The recorded finalisation time of the GFS run advanced 77 times over the following 100 minutes, ending 5.3 hours after its issue time. This matches the median finalisation delay of 5.22 hours observed in the full archive. The archive retains only the final values of each run; the intermediate states visible before finalisation are not kept. The publication delays are also consistent with the weather centres' own release schedules. ECMWF disseminates the first 90 hours of its 00 UTC high-resolution (control) run from 05:45 UTC and of its 00 UTC ensemble from 06:40 UTC (ECMWF, 2026); the median publication delays we observe for the corresponding configurations are 5.78 and 6.73 hours.

\subsection{The information set at a gate}

Let $g_d$ denote the gate closure, in UTC, of a given auction for delivery day $d$, and let $\Delta \ge 0$ denote an \emph{action margin}: the time a participant needs between receiving information and submitting orders. The information set available at the decision point is
\[\mathcal{I}_d(\Delta) = \{\, \text{information available at or before } g_d - \Delta \,\}.\]

A forecast run $r$ is admissible at the decision point if it belongs to $\mathcal{I}_d(\Delta)$. Two conditions are natural. The run is \emph{published} if $\tau^{P}_r \le g_d - \Delta$. It is \emph{final} if $\tau^{F}_r \le g_d - \Delta$. Only the second condition guarantees that the values stored in the archive, which are the values after finalisation, coincide with the values a participant could have seen at the decision point.

The two conditions separate two mechanisms by which a dataset can contain information that was not available at the decision point. \emph{Run-availability leakage} occurs when the selected run had not been published at all. \emph{Value-revision exposure} occurs when the run had been published but its values changed afterwards, so that the archived values may differ from those visible at the decision point. The first is a definite violation. The second is a risk whose size cannot be measured from the archive, because intermediate values are not kept. Three information sets should accordingly be distinguished: the information actually observable at the decision point, $\mathcal{I}^{\text{actual}}$; the set reconstructed from the archive by the publication condition, $\mathcal{I}^{\text{archive}}$, which contains the right runs but possibly revised values; and the set reconstructed by the finalisation condition, $\mathcal{I}^{\text{cons}}$, a reproducible subset of $\mathcal{I}^{\text{actual}}$ that may exclude runs a participant could in fact have used. Historical reproducibility and operational availability therefore do not coincide.

\subsection{Three selection rules}

A point-in-time dataset assigns to each delivery day and gate one forecast run per weather model, typically the most recent admissible one. We compare three selection rules. For a weather model with runs $\mathcal{R}$:

\textbf{Naive rule.} Select the most recently issued run whose issue time precedes the gate closure, with the gate expressed in a fixed UTC convention $\tilde{g}_d$:

\[r^{\text{naive}}_d = \arg\max_{r \in \mathcal{R}} \{\tau^{I}_r : \tau^{I}_r \le \tilde{g}_d\}.\]

\textbf{Publication rule.} Select the most recently issued run that was published by the true gate closure less the action margin:

\[r^{\text{pub}}_d(\Delta) = \arg\max_{r \in \mathcal{R}} \{\tau^{I}_r : \tau^{P}_r \le g_d - \Delta\}.\]

\textbf{Finalisation rule.} Select the most recently issued run that was final by the true gate closure less the action margin:

\[r^{\text{fin}}_d(\Delta) = \arg\max_{r \in \mathcal{R}} \{\tau^{I}_r : \tau^{F}_r \le g_d - \Delta\}.\]

The naive rule is the one that data interfaces make easiest. Runs are indexed and queried by issue time, and requesting ``the latest run issued before the gate'' is the obvious query. It treats the issue time as if it were the publication time (Pitfall 1). A second, independent error is to state the gate in a fixed UTC convention when the auction runs on local clock time (Pitfall 2). Because runs are issued on whole cycle hours, this error rarely changes the run chosen by the naive rule. It does, however, defeat the publication and finalisation rules: with the gate stated in UTC, these rules accept, in summer, runs that become available in the hour after the true gate. We therefore always evaluate the two corrected rules at the true gate $g_d$. The publication rule removes run-availability leakage but remains exposed to value revision. The finalisation rule removes both: under it, the archived values coincide with values available at the decision point. It is conservative rather than exact. Finalisation is recorded per run, not per forecast step, so a run whose delivery-day values had already settled before the gate is still rejected. Where historical as-of values are unavailable, run-level finalisation provides a conservative, auditable selection rule, at the cost of potentially excluding information available in real time. Figure 2 illustrates the three rules for a single gate.

\begin{figure*}[tp]
\centering
\includegraphics[width=\textwidth]{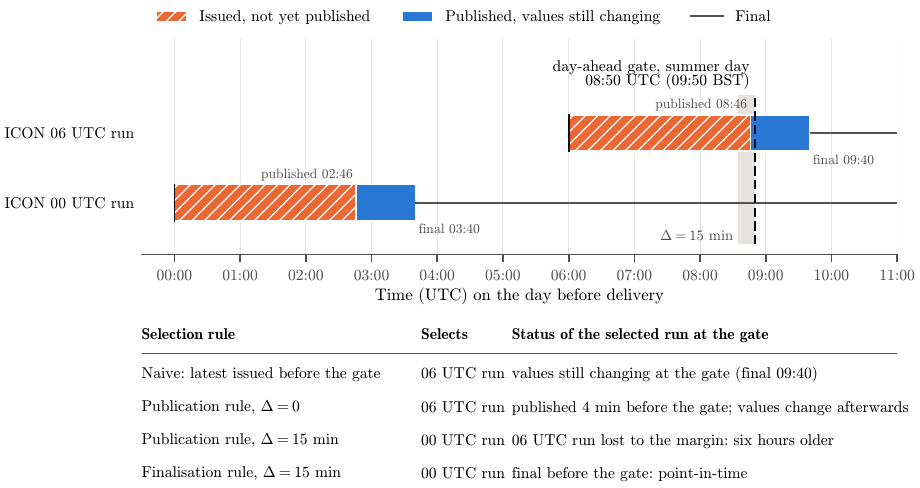}
\caption{Selection rules at one gate, illustrated for ICON at the summer day-ahead gate (09:50 BST, 08:50 UTC), using the configuration's median publication and finalisation delays. The 06 UTC run is published four minutes before the gate and finalised fifty minutes after it. The naive rule and the publication rule without an action margin both select the 06 UTC run, whose archived (final) values were not those visible at the gate. With an action margin of 15 minutes (shaded), the publication rule falls back to the 00 UTC run, six hours older. The finalisation rule selects the 00 UTC run, whose values were final before the gate. All rules are evaluated at the true gate; the effect of stating the gate in UTC (Pitfall 2) is shown in Figure 1 and Table 4.}
\end{figure*}

\section{Data}

We use the commercial forecast archive introduced in Section 3.2. It contains forecasts of three variables for Great Britain: wind power generation, solar photovoltaic generation, and electricity demand (consumption). Each variable is forecast from ten weather-model configurations (Table 1). For the audit in Sections 5.1--5.4 we retrieved only the metadata of each run, namely the three timestamps defined in Section 3.2. The retrieval covers runs issued from 25 December 2021 to 1 October 2026, which spans delivery days from 1 January 2022 to 30 September 2026 (1,734 days). In total we obtained 128,656 runs. Two configurations (AIFS and AIFS-ENS) are available only from 2026 onwards. For 3,822 runs (3.0\%) no finalisation time is recorded; under the finalisation rule we treat such runs as never final, which is the conservative choice.

For the comparison of forecast values in Section 5.5 we also use the values of the runs selected at the day-ahead and IDA1 gates, for the 882 delivery days from 1 January 2024 to 31 May 2026: the naive-rule selections, archived by an earlier version of our own pipeline in June 2026, and 10,932 runs selected by the corrected rules, retrieved in October 2026. Re-retrieving 36 of the archived runs in October 2026 reproduced their values exactly, so the two retrievals are comparable. Outturns of wind generation, solar generation and demand, at half-hourly resolution, are taken from the same vendor.

\begin{table*}[tp]
\centering\small
\caption{Weather-model configurations in the archive.}
\begin{tabularx}{\textwidth}{l>{\raggedright\arraybackslash\hsize=1.00\hsize}Xlll}
\toprule
\textbf{Configuration} & \textbf{Underlying system} & \textbf{Cycles (UTC)} & \textbf{Horizon} & \textbf{Runs per variable} \\
\midrule
GFS & NCEP Global Forecast System & 00, 06, 12, 18 & up to 16 days & 6,957 \\
GFS-ENS & NCEP global ensemble & 00, 06, 12, 18 & n/a & 6,952 \\
ICON & DWD ICON global, main runs & 00, 06, 12, 18 & 120 h & 6,955 \\
ICON-SR & DWD ICON global, short-range intermediate runs & 03, 09, 15, 21 & 30 h & 6,945 \\
ECMWF & ECMWF IFS high-resolution, main runs & 00, 12 & 360 h & 3,482 \\
ECMWF-ENS & ECMWF IFS ensemble, main runs & 00, 12 & 360 h & 3,483 \\
ECMWF-SR & ECMWF IFS high-resolution, 06/18 runs & 06, 18 & 144 h & 3,480 \\
ECMWF-SR-ENS & ECMWF IFS ensemble, 06/18 runs & 06, 18 & 144 h & 3,476 \\
AIFS & ECMWF AI forecasting system & 00, 06, 12, 18 & 360 h & 640 (2026) \\
AIFS-ENS & ECMWF AI ensemble & 00, 06, 12, 18 & n/a & 512 (2026) \\
\bottomrule
\end{tabularx}
\par\smallskip\begin{minipage}{\textwidth}\footnotesize Runs per variable: wind generation; counts for solar and demand differ by at most one run. Horizon: forecast range available in the archive, measured on runs issued in October 2026 (for GFS the range varies by cycle; GFS-ENS and AIFS-ENS were not checked). It can be shorter than the weather centre's own range. The mapping of configurations to weather-model runs follows from their cycle times and horizons, which match the centres' documented schedules. DWD runs ICON main forecasts at 00, 06, 12 and 18 UTC and ``additional short-range forecasts starting at 03, 09, 15, 21 UTC'' (Reinert et al., 2026, §2.3.2); ECMWF caps its 06 and 18 UTC runs at step 144 (ECMWF, 2026).\end{minipage}
\end{table*}

Gate closures are taken from the exchanges' published timetables (Section 3.1) and converted from UK clock time to UTC for every delivery day, so the one-hour shift at each daylight-saving transition is included. Of the 1,734 delivery days, 1,039 have their D$-$1 gate closures in BST and 695 in GMT.

\section{Results}

\subsection{Publication delays are long, model-specific and not constant}

Table 2 reports the delay between issue and publication, and between issue and finalisation, for each configuration. The delays are long relative to the spacing of the auction gates. The fastest configurations (ICON and ICON-SR) are published about 2.8 hours after their issue time. GFS follows at 3.6 hours, and the ECMWF configurations at 5.4 to 6.7 hours. Finalisation adds up to a further 2.7 hours (GFS-ENS).

\begin{table*}[tp]
\centering\small
\caption{Delays after issue time (hours), wind generation, 2022--2026, runs with both times recorded.}
\begin{tabularx}{\textwidth}{l>{\raggedright\arraybackslash\hsize=1.00\hsize}X>{\raggedright\arraybackslash\hsize=1.00\hsize}X>{\raggedright\arraybackslash\hsize=1.00\hsize}X>{\raggedright\arraybackslash\hsize=1.00\hsize}X>{\raggedright\arraybackslash\hsize=1.00\hsize}X}
\toprule
\textbf{Configuration} & \textbf{Publication: median} & \textbf{Publication: 95th pct.} & \textbf{Finalisation: median} & \textbf{Finalisation: 95th pct.} & \textbf{No finalisation time (\%)} \\
\midrule
ICON-SR & 2.75 & 3.00 & 3.20 & 3.30 & 3.3 \\
ICON & 2.76 & 3.09 & 3.67 & 3.83 & 0.4 \\
GFS & 3.55 & 3.64 & 5.22 & 5.37 & 0.1 \\
GFS-ENS & 3.86 & 3.94 & 6.52 & 6.75 & 0.3 \\
AIFS (2026) & 5.42 & 5.80 & 5.48 & 5.88 & 8.4 \\
ECMWF-SR & 5.77 & 5.78 & 6.21 & 6.22 & 6.9 \\
ECMWF & 5.78 & 5.79 & 6.93 & 6.97 & 6.9 \\
AIFS-ENS (2026) & 5.84 & 6.24 & 5.92 & 6.51 & 3.9 \\
ECMWF-SR-ENS & 6.71 & 6.76 & 7.08 & 7.41 & 6.8 \\
ECMWF-ENS & 6.72 & 6.85 & 7.69 & 8.30 & 6.0 \\
\bottomrule
\end{tabularx}
\par\smallskip\begin{minipage}{\textwidth}\footnotesize Delays are computed on the runs for which both a publication and a finalisation time are recorded, so that the four delay columns describe the same runs. The last column gives the share of all runs without a recorded finalisation time; every run has a publication time.\end{minipage}
\end{table*}

Within each configuration the delays are tightly concentrated: for six of the eight configurations with a full history, the 95th percentile of the publication delay lies within 0.15 hours of the median. The delays are, however, not constant over time. The median publication delay of the four ECMWF configurations was stable from 2022 until mid-2024 (5.77--6.76 hours) and, after a transition in the second half of 2024, fell by between 0.6 and 1.4 hours (2025 medians 5.14--5.65 hours). Delays can also lengthen. The median publication delay of the ICON 06 UTC run rose steadily from 2.68 hours (late 2021) to 2.83 hours (second half of 2025) and 3.20 hours (second half of 2026); at about 2.83 hours the run reaches the summer day-ahead gate (08:50 UTC), which it has missed on almost every summer day since 2026 (Figure 3). Any rule that assumes a fixed publication delay is therefore exposed to silent changes in the vendor's processing chain.

Finalisation times are missing for 3.0\% of runs, and the gaps are concentrated in three episodes: the four ECMWF configurations between May and September 2024, during the transition described above (27--32\% of their 2024 runs, and 81--100\% of those issued from June to August 2024), ICON-SR in 2026 (17.8\%) and AIFS (8.4\%). Elsewhere the share is below 1\% per year. Under the finalisation rule such runs are treated as never final, so during the 2024 episode the rule falls back to ECMWF runs several days old.

Drift is not an error in dataset construction but a change in the environment, and it turns otherwise reasonable practices into pitfalls. A rule that assumes a fixed delay leaks silently when delays lengthen and discards usable runs when they shorten; thresholds calibrated once (Section 5.4) shift one-for-one with the delay; and a leakage rate measured over one period need not hold in another (Section 5.3). Drift also affects datasets that are built correctly. Under the publication rule, ICON at the summer day-ahead gate supplied its 06 UTC run, about 2.8 hours old, in 2022--2025, but its 00 UTC run, about 8.8 hours old, in 2026. A model trained on the earlier period meets systematically older inputs in the later one, although both datasets are point-in-time correct.

\begin{figure*}[tp]
\centering
\includegraphics[width=\textwidth]{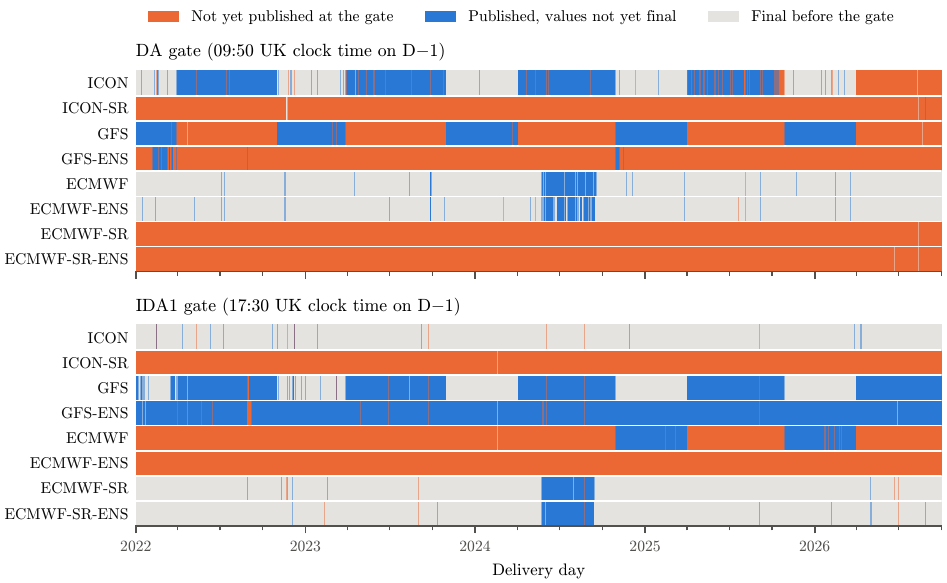}
\caption{Status at the true gate closure of the run selected by the naive rule, for every delivery day from January 2022 to September 2026 (columns) and each weather-model configuration (rows), at the day-ahead gate (top) and the IDA1 gate (bottom). Orange: the selected run had not yet been published. Blue: it had been published but its values were not yet final. Grey: it was final before the gate. Wind generation; solar and demand give the same picture. AIFS configurations, available only from 2026, are omitted. Summer stripes (GFS and ICON at the day-ahead gate, GFS at the IDA1 gate) reflect the daylight-saving shift of the gates. Changes of regime reflect drifting publication delays: ECMWF at the IDA1 gate in winter from late 2024, after its delays shortened, and ICON at the day-ahead gate in summer 2026, after its delay lengthened past the gate. The block in mid-2024 marks ECMWF runs without a recorded finalisation time (May to September 2024), which are counted as not final.}
\end{figure*}

\subsection{Pitfall 1: the issue time is not the publication time}

Table 3 reports, for each configuration and gate, the share of delivery days on which the run selected by the naive rule had not yet been published at the true gate closure, together with the share on which it had not yet been finalised.

\begin{table*}[tp]
\centering\small
\caption{Naive-rule leakage and exposure: share of delivery days (\%) on which the selected run was not yet published (run-availability leakage), or not yet final (value-revision exposure), at the true gate closure. Wind generation, 2022--2026.}
\begin{tabularx}{\textwidth}{l>{\raggedright\arraybackslash\hsize=1.00\hsize}X>{\raggedright\arraybackslash\hsize=1.00\hsize}X>{\raggedright\arraybackslash\hsize=1.00\hsize}X>{\raggedright\arraybackslash\hsize=1.00\hsize}X}
\toprule
\textbf{Configuration} & \textbf{DA, not published (BST / GMT)} & \textbf{DA, not final (BST / GMT)} & \textbf{IDA1, not published (BST / GMT)} & \textbf{IDA1, not final (BST / GMT)} \\
\midrule
GFS & 99.8 / 0.6 & 99.8 / 100.0 & 0.9 / 0.1 & 99.7 / 5.0 \\
GFS-ENS & 99.8 / 93.2 & 100.0 / 100.0 & 2.3 / 0.0 & 99.8 / 99.6 \\
ICON & 23.5 / 1.0 & 99.8 / 4.0 & 0.5 / 0.6 & 1.3 / 1.4 \\
ICON-SR & 99.7 / 99.6 & 99.8 / 99.6 & 100.0 / 99.9 & 100.0 / 99.9 \\
ECMWF & 0.1 / 0.0 & 11.5 / 1.2 & 100.0 / 57.0 & 100.0 / 99.9 \\
ECMWF-ENS & 0.3 / 0.1 & 9.9 / 1.2 & 100.0 / 100.0 & 100.0 / 100.0 \\
ECMWF-SR & 99.9 / 100.0 & 99.9 / 100.0 & 0.5 / 0.1 & 11.5 / 0.7 \\
ECMWF-SR-ENS & 99.8 / 100.0 & 99.8 / 100.0 & 0.4 / 0.1 & 11.5 / 0.4 \\
AIFS (2026) & 99.4 / n/a & 99.4 / n/a & 100.0 / n/a & 100.0 / n/a \\
AIFS-ENS (2026) & 99.2 / n/a & 99.2 / n/a & 100.0 / n/a & 100.0 / n/a \\
\bottomrule
\end{tabularx}
\par\smallskip\begin{minipage}{\textwidth}\footnotesize DA: N2EX day-ahead gate (09:50 UK time, D$-$1). IDA1: EPEX first intraday auction gate (17:30 UK time, D$-$1). BST/GMT refers to the clock regime in force at the gate.\end{minipage}
\end{table*}

The naive rule frequently selects a run that did not exist at the gate. At the day-ahead gate, it does so on essentially every day for ICON-SR and both ECMWF short-range configurations, regardless of season. In each case the most recent cycle precedes the gate by less than the configuration's publication delay. At the intraday gate the same is true of ICON-SR, both ECMWF global configurations and both AIFS configurations. Even where the selected run had been published, its values were often not yet final. GFS at the day-ahead gate is the clearest example: its 06 UTC run is published at about 09:33 UTC, before the winter gate (09:50 UTC), but is finalised only at about 11:13 UTC. The naive rule is therefore exposed to value revision on every day of the sample: the archived values it supplies are not necessarily those visible at the gate.

The leakage is a property of the configuration and the gate, not of the forecast variable. Repeating the analysis for solar generation and demand changes no entry in Table 3 by more than 0.1 percentage points. This is expected, since the vendor publishes the three variables from the same weather-model runs. It means that the leakage of any point-in-time dataset built from such an archive can be determined from timestamps alone, before any model is fitted.

\subsection{Pitfall 2: gates follow local clock time}

In winter, UK clock time coincides with UTC. In summer, every gate closure falls one hour earlier in UTC, while the weather models keep their UTC cycle times and the vendor's publication delays stay the same. Two consequences follow.

First, the leakage of the naive rule depends strongly on the season (Table 3). For GFS at the day-ahead gate, the naive selection had not been published at the gate on 0.6\% of winter days but on 99.8\% of summer days. For ICON at the same gate the figures are 1.0\% and 23.5\%. In the 2024--2026 dataset used by an earlier version of our own pipeline (GFS for wind and demand, ICON for solar), the share of day-ahead selections that were unpublished at the gate was 0.3\% in winter and 74.1\% in summer. A dataset checked only on winter days would therefore appear almost clean.

Second, stating the gate in UTC defeats rules that are otherwise correct. Table 4 applies the publication and finalisation rules with the gate in a fixed UTC convention, and reports how often the selected run was not available at the true gate on summer days. A publication rule with a UTC gate selects an unpublished GFS run at the day-ahead gate on 98.9\% of summer days, and an unpublished ECMWF run at the intraday gate on 35.6\%. A finalisation rule with a UTC gate selects a non-final ICON run at the day-ahead gate on 94.6\% of summer days, and a non-final GFS run at the intraday gate on 96.3\%. In other words, using publication or finalisation times is not sufficient: the gate must also be stated in the market's clock time.

\begin{table*}[tp]
\centering\small
\caption{Leakage caused by stating the gate in UTC alone: share of summer (BST) delivery days (\%) on which the run selected with a fixed-UTC gate was not available at the true gate. Wind generation, 2022--2026, $\Delta$ = 0. Configurations with no entry above 1\% are omitted.}
\begin{tabularx}{\textwidth}{ll>{\raggedright\arraybackslash\hsize=1.00\hsize}X>{\raggedright\arraybackslash\hsize=1.00\hsize}X}
\toprule
\textbf{Gate} & \textbf{Configuration} & \textbf{Publication rule, UTC gate: not published} & \textbf{Finalisation rule, UTC gate: not final} \\
\midrule
DA & GFS & 98.9 & 0.4 \\
DA & GFS-ENS & 11.4 & 0.5 \\
DA & ICON & 23.3 & 94.6 \\
IDA1 & ECMWF & 35.6 & 0.0 \\
IDA1 & GFS & 0.3 & 96.3 \\
\bottomrule
\end{tabularx}
\end{table*}

Daylight saving also interacts with the drift reported in Section 5.1. Before 2025, the ECMWF 12 UTC run was published at about 17:47 UTC, after the intraday gate in both seasons. After its publication delay shortened from late 2024, the same run became available before the winter gate (17:30 UTC) but remained unavailable before the summer gate (16:30 UTC). The resulting naive-rule leakage at the winter intraday gate, 57.0\%, is an average over two regimes and describes neither of them (Figure 3).

\begin{figure*}[tp]
\centering
\includegraphics[width=\textwidth]{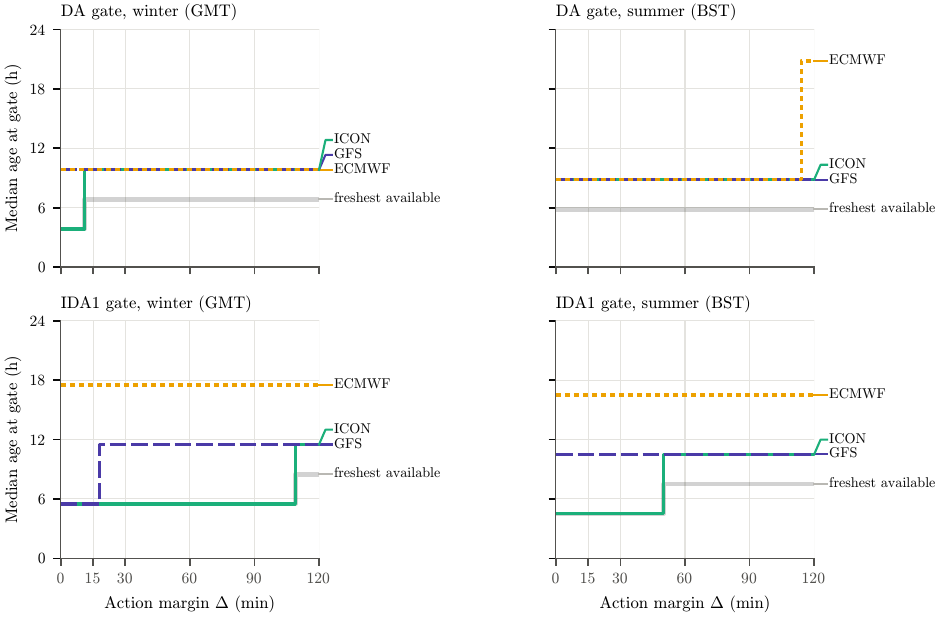}
\caption{Median age at the gate of the run selected by the finalisation rule, as a function of the action margin $\Delta$, for the day-ahead gate (top) and the IDA1 gate (bottom), in winter (left) and summer (right). Coloured lines: ICON, GFS and ECMWF. Grey band: the freshest admissible run across all eight configurations with a full history. Wind generation, delivery days January 2022 to September 2026. Each step marks the margin at which the most recent run is lost and the selection falls back to the previous cycle.}
\end{figure*}

\subsection{The action margin: freshness is lost in steps}

Under the publication and finalisation rules, the action margin $\Delta$ determines which run is selected. We evaluated both rules for every whole minute $\Delta \in \{0, 1, \dots, 120\}$ and recorded the median age of the selected run at the gate, $g_d - \tau^{I}_{r}$, separately for winter and summer gates. Figure 4 shows the results for the finalisation rule; Table 5 lists every step for both rules.

For most combinations of configuration, gate and season, $\Delta$ has no effect anywhere in the range 0--120 minutes: the selected run, and therefore its age, is the same at every margin. Where $\Delta$ does matter, its effect is a single step. Once the margin exceeds the time between a run's availability and the gate, that run is lost, and the selection falls back to the previous cycle. The cost is 6 hours of forecast age for configurations with four daily cycles, and 12 hours for configurations with two (Table 5).

\begin{table*}[tp]
\centering\small
\caption{Every combination for which the selected run changes within $\Delta$ $\le$ 120 minutes (median age of the selected run at the gate, hours; thresholds at one-minute resolution).}
\begin{tabularx}{\textwidth}{l>{\raggedright\arraybackslash\hsize=1.00\hsize}X>{\raggedright\arraybackslash\hsize=1.00\hsize}X>{\raggedright\arraybackslash\hsize=1.00\hsize}X>{\raggedright\arraybackslash\hsize=1.00\hsize}X>{\raggedright\arraybackslash\hsize=1.00\hsize}X>{\raggedright\arraybackslash\hsize=1.00\hsize}X}
\toprule
\textbf{Gate} & \textbf{Configuration} & \textbf{Rule} & \textbf{Season} & \textbf{Age at $\Delta$ = 0} & \textbf{Step at $\Delta$ (min)} & \textbf{Age after step} \\
\midrule
DA & ICON & publication & BST & 2.8 & 5 & 8.8 \\
DA & ICON & finalisation & GMT & 3.8 & 11 & 9.8 \\
DA & GFS & publication & GMT & 3.8 & 19 & 9.8 \\
DA & ICON & publication & GMT & 3.8 & 64 & 9.8 \\
DA & ECMWF-ENS & finalisation & BST & 8.8 & 69 & 20.8 \\
DA & ECMWF & finalisation & BST & 8.8 & 114 & 20.8 \\
IDA1 & GFS & finalisation & GMT & 5.5 & 18 & 11.5 \\
IDA1 & GFS-ENS & publication & BST & 4.5 & 39 & 10.5 \\
IDA1 & ICON & finalisation & BST & 4.5 & 50 & 10.5 \\
IDA1 & GFS & publication & BST & 4.5 & 57 & 10.5 \\
IDA1 & GFS-ENS & publication & GMT & 5.5 & 99 & 11.5 \\
IDA1 & ICON & publication & BST & 4.5 & 105 & 10.5 \\
IDA1 & ICON & finalisation & GMT & 5.5 & 109 & 11.5 \\
IDA1 & GFS & publication & GMT & 5.5 & 118 & 11.5 \\
\bottomrule
\end{tabularx}
\par\smallskip\begin{minipage}{\textwidth}\footnotesize ICON-SR, ECMWF-SR and ECMWF-SR-ENS show no step within the range. AIFS configurations are omitted because their history is too short.\end{minipage}
\end{table*}

Several of these thresholds are short. Under the publication rule, the ICON 06 UTC run becomes available only minutes before the summer day-ahead gate, and a margin of 5 minutes is enough to lose it. Under the finalisation rule, the same run is lost at the winter day-ahead gate at a margin of 11 minutes, and the GFS 12 UTC run at the winter intraday gate at 18 minutes. A margin of only a few minutes for processing a forecast and submitting orders can therefore remove the freshest run, without any visible sign in the resulting dataset.

Because the age of the selected run cannot decrease as $\Delta$ grows, these results are a lower bound on staleness for participants who need longer margins. With regular cycles of length $C$ and a stable delay $d$ (publication or finalisation), the age of the run selected at margin $\Delta$ satisfies $\Delta + d \le \text{age} < \Delta + d + C$: beyond the first threshold, staleness grows roughly hour for hour with the margin.

The action margin also changes which configuration is the most current. At the winter day-ahead gate under the finalisation rule, ICON provides a forecast 3.8 hours old at $\Delta = 0$, but at $\Delta \ge 11$ minutes no configuration is younger than 6.8 hours (ICON-SR), and no other configuration is younger than 9.8 hours. The naive rule, finally, overstates the freshness of the information by a full cycle. At the day-ahead gate it assigns GFS forecasts an age of 2.8--3.8 hours, whereas under the finalisation rule the youngest admissible GFS run is 8.8--9.8 hours old.

\subsection{How different are the leaked inputs?}

Sections 5.2--5.4 measure how often a selection rule uses information that was not available, not how much that information differs from what was. We therefore compared the forecast values selected by the naive rule with those selected by the two corrected rules, for wind generation, solar generation and demand, three deterministic configurations (GFS, ICON, and ECMWF, using ECMWF-SR at the day-ahead gate and ECMWF at the IDA1 gate), both gates and the 882 delivery days from January 2024 to May 2026. The corrected rules were evaluated at the true gate with $\Delta = 0$, the most favourable case for them. Forecasts were averaged to half-hours and compared with the outturns.

\begin{table*}[tp]
\centering\small
\caption{Naive rule versus finalisation rule: difference between the selected inputs and apparent accuracy gain of the naive selection, delivery days 2024--2026.}
\begin{tabularx}{\textwidth}{lll>{\raggedright\arraybackslash\hsize=0.91\hsize}X>{\raggedright\arraybackslash\hsize=0.91\hsize}X>{\raggedright\arraybackslash\hsize=1.17\hsize}Xl}
\toprule
\textbf{Variable} & \textbf{Configuration} & \textbf{Gate} & \textbf{Days with a different run (\%)} & \textbf{Mean $D$, MW (\% of mean outturn)} & \textbf{$G$, all days (\%) [95\% CI]} & \textbf{$G$, summer (\%)} \\
\midrule
Wind & ECMWF-SR & DA & 100 & 611 (6.1) & 11.4 [8.9, 14.0] & 9.2 \\
Wind & ECMWF & IDA1 & 100 & 550 (5.5) & 10.1 [7.9, 12.2] & 9.0 \\
Wind & GFS & DA & 100 & 486 (4.9) & 5.4 [3.7, 7.0] & 4.9 \\
Wind & GFS & IDA1 & 55 & 240 (2.4) & 4.2 [3.0, 5.6] & 7.9 \\
Wind & ICON & DA & 56 & 305 (3.1) & 0.0 [$-$1.4, 1.2] & $-$0.2 \\
Solar & ECMWF-SR & DA & 100 & 126 (6.6) & 2.7 [0.4, 5.1] & 1.0 \\
Solar & ECMWF & IDA1 & 100 & 125 (6.5) & 3.0 [1.1, 5.0] & 2.2 \\
Solar & GFS & DA & 100 & 117 (6.1) & 0.6 [$-$1.0, 2.2] & $-$0.3 \\
Solar & GFS & IDA1 & 55 & 87 (4.5) & 2.4 [1.2, 3.5] & 3.3 \\
Solar & ICON & DA & 56 & 74 (3.8) & 1.7 [0.2, 3.1] & 2.5 \\
Demand & ECMWF-SR & DA & 100 & 381 (1.3) & 13.0 [11.5, 14.5] & 17.1 \\
Demand & ECMWF & IDA1 & 100 & 478 (1.6) & 12.4 [10.8, 14.0] & 16.6 \\
Demand & GFS & DA & 100 & 397 (1.3) & 17.9 [16.2, 19.7] & 24.0 \\
Demand & GFS & IDA1 & 55 & 120 (0.4) & $-$4.6 [$-$5.8, $-$3.4] & $-$8.6 \\
Demand & ICON & DA & 56 & 264 (0.9) & 13.5 [11.3, 15.9] & 24.2 \\
\bottomrule
\end{tabularx}
\par\smallskip\begin{minipage}{\textwidth}\footnotesize Days with a different run: share of delivery days on which the finalisation rule selects a different run from the naive rule; $D$ is averaged over all days. ICON at the IDA1 gate is omitted (different run on about 1\% of days). ECMWF rows exclude about 110 days in 2024 on which no ECMWF run with a recorded finalisation time covered the delivery day (Section 5.1); they cover 771 (day-ahead) and 790 (IDA1) days.\end{minipage}
\end{table*}

For delivery day $d$, let $H_d$ be the half-hours at which all forecasts and the outturn $y$ are available, and let $F^{\text{naive}}$ and $F^{\text{rule}}$ be the values of the runs selected by the naive rule and by a corrected rule. We report the mean absolute difference between the two selections and the mean absolute error of each,
\[D_d = \frac{1}{|H_d|}\sum_{h \in H_d} \left|F^{\text{naive}}(h) - F^{\text{rule}}(h)\right|,\]
\[\text{MAE}^{k}_d = \frac{1}{|H_d|}\sum_{h \in H_d} \left|F^{k}(h) - y(h)\right|,\]

and the \emph{apparent accuracy gain} of the naive selection,
\[G = \frac{\overline{\text{MAE}}^{\text{rule}} - \overline{\text{MAE}}^{\text{naive}}}{\overline{\text{MAE}}^{\text{rule}}},\]

where bars denote averages over days. $D$ measures how different the inputs are; $G$ measures how much more accurate the naively selected inputs appear than the inputs available at the gate. A positive $G$ means that the leakage makes the inputs look more accurate than those available at the gate. Confidence intervals for $G$ are obtained by a paired, circular moving-block bootstrap over delivery days (Künsch, 1989; Politis and Romano, 1992), with blocks of 7 consecutive days, 2,000 resamples and percentile intervals. Resampling blocks preserves the day-to-day dependence of weather-driven errors, and resampling both selections on the same days preserves the pairing. The intervals are essentially unchanged for block lengths from 1 to 28 days, and no conclusion depends on the choice.

Two results stand out (Table 6). First, wherever the two rules select different runs, the inputs differ materially: by 2--7\% of mean outturn for wind and solar, several hundred megawatts for wind, and 1--2\% for demand. The naive rule changes the numbers a model sees, not only the label of the run. Second, the effect on apparent accuracy varies widely and is not always positive. It is large for demand at the day-ahead gate (13--18\%, and up to 24\% in summer) and for ECMWF wind (10--11\%), moderate for GFS wind (4--8\%) and small for solar (at most 3\%). For ICON wind at the day-ahead gate it is nil, although the selected runs differ on 99.8\% of summer days, and for GFS demand at the IDA1 gate it is negative: in summer the naively selected run is 8.6\% \emph{less} accurate. Under the publication rule, the differences for GFS and ICON are confined almost entirely to summer, in line with Section 5.3, whereas ECMWF differs on every day at the day-ahead gate in both seasons. Removing each rule's mean error before computing $G$ changes it by at most one percentage point for wind and demand, so constant offsets between the forecast and outturn series do not drive these results; for solar, part of the small ECMWF effect in winter reflects such an offset.

The demand results are not explained by forecast age. Using, for each delivery day, the run selected by the finalisation rule at the day-ahead gate one to five days earlier shows how accuracy declines with lead time: a run one day older is 25--35\% less accurate for wind, about 10\% for solar, but only 1.5--3\% for demand. A run six hours older should therefore cost demand well under 1\%, not 18\%. Scoring every run issued in the two days before delivery against the same delivery day resolves the puzzle. For wind, accuracy improves steadily from each run to the next. For demand, it follows a daily cycle that repeats in all three configurations: the 06 UTC run is the most accurate of the day, accuracy declines through the 12 and 18 UTC runs, and the 00 UTC run is the least accurate, worse even than the 18 UTC run issued six hours earlier (GFS mean absolute error 751, 787, 851 and 920 MW for the 06, 12, 18 and 00 UTC runs of the day before delivery). Since the pattern is common to three weather models, it suggests a shared stage of the vendor's demand processing; its cause is not established. At the day-ahead gate the naive rule happens to select the most accurate demand run of the day and the finalisation rule the least accurate one; at the IDA1 gate in summer the roles are reversed. Which run a rule admits can thus matter more than how old it is. A model trained on naively selected data would therefore be trained on systematically different, and at the day-ahead gate more accurate, forecast inputs than those available under point-in-time-correct selection; the consequences for downstream predictive performance remain to be measured.

The lead-time results also put a price on the steps of Section 5.4. Losing one six-hour cycle costs little accuracy for demand, but for wind it means a run roughly a quarter of a day older, and a correspondingly less accurate forecast.

These comparisons have limits. They compare archived values, so they measure how different the wrongly selected inputs are, not the exact values visible at the gate, which the archive does not retain. The outturns used as reference are themselves revised (Section 6). They use $\Delta = 0$; longer margins select older runs, and how that changes these differences requires separate evaluation. The comparison covers 2024--2026 and three deterministic configurations only.

\section{Beyond forecasts: point-in-time exposure across data types}

Forecast runs are the most exposed input to point-in-time errors, but they are not the only one. Models of day-ahead and intraday prices typically also use auction prices and outturn series, such as metered or estimated generation and demand, as features or targets. We examined both types for the same market, using two complementary checks. First, we compared the latest available value of each series with the time of the request, which gives the publication delay. Second, we compared a copy of each series retrieved on 1 June 2026 with a fresh retrieval of the same period (January 2024 to May 2026) on 7 October 2026. Any difference between the two copies is a revision made after the first retrieval.

\textbf{Auction prices.} Neither the N2EX day-ahead price nor the EPEX IDA1 price changed between the two retrievals (0 of 21,143 hourly and 0 of 42,190 half-hourly values). Auction prices become known at a precise, documented time (10:00 and 18:00 UK time on D$-$1 respectively), and they can be used as features for any decision taken after that time. The IDA1 price is unknown at the IDA1 gate. It may be a prediction target at that gate but never a feature.

\textbf{Outturn series.} Generation and demand outturns are published close to real time. At the time of our check, the latest value of wind generation referred to a half-hour that had ended about 1.3 hours earlier, and the latest values of solar generation and demand to half-hours that had ended about 20 minutes earlier. Their values, however, are revised after publication. Between the two retrievals, wind generation changed on every delivery day from 30 April to 31 May 2026, that is, the last 32 days before the first retrieval, with a median relative change of 0.9\% among changed values. Demand changed over the same window by a median of 0.2--0.3\%. Solar generation changed sporadically, on 13 days in May, with a median relative change of 4.4\%. No value older than the 32-day window changed. The comparison therefore identifies revisions affecting observations up to 32 days old at the first retrieval. This is consistent with an approximately monthly revision process, but does not establish a definitive revision window: earlier or reversed revisions would not be visible.

This revision behaviour cannot be fully audited from a vendor archive, for a simple reason. Unlike forecast runs, outturn series are stored as a single, continuously overwritten series, so earlier values are not retained. Our comparison shows only the revisions made after our first retrieval, and is therefore a lower bound. Revisions made in the first hours after publication are invisible. The vendor does not document the upstream source of the outturn series, so the revision cycle cannot be checked against the source's own policy either.

Table 7 summarises the point-in-time properties of the three data types.

\begin{table*}[tp]
\centering\small
\caption{Point-in-time exposure by data type (GB, 2022--2026).}
\begin{tabularx}{\textwidth}{l>{\raggedright\arraybackslash\hsize=0.93\hsize}X>{\raggedright\arraybackslash\hsize=0.82\hsize}X>{\raggedright\arraybackslash\hsize=0.64\hsize}X>{\raggedright\arraybackslash\hsize=1.62\hsize}X}
\toprule
\textbf{Data type} & \textbf{When known} & \textbf{Revised after publication?} & \textbf{Auditable from the archive?} & \textbf{Safe use} \\
\midrule
Forecast runs & Publication time, 2.8--6.7 h after issue (Table 2) & Yes, until finalisation, up to 2.7 h later & Yes: issue, publication and finalisation times stored per run & Select by the finalisation rule with an explicit margin $\Delta$ (Section 7) \\
Auction prices & At results publication (DA 10:00, IDA1 18:00 UK time, D$-$1) & None observed (two retrievals) & Not tested & As a feature only after publication; never as a feature at its own gate \\
Outturns & Close to real time ($\approx$ 0.3--1.3 h after the half-hour) & Yes; observed up to 32 days after delivery (median changes 0.2--4.4\%) & No: only the latest values are retained & Lag conservatively (revisions seen up to 32 days), or record first-published values going forward; otherwise declare the revision exposure \\
\bottomrule
\end{tabularx}
\end{table*}

\section{Recommended protocol}

The three pitfalls share one cause: the past is reconstructed from archives that do not fully preserve what was observable at each decision time. This suggests a principle for building datasets: \emph{audit the past, record the future}. Historical data can only be audited, from whatever timestamps the archive provides. Going forward, every version of every input can be recorded when it arrives, together with its arrival time; a dataset recorded in this way can be replayed exactly as of any decision time, margin or gate. The protocol below implements both halves. None of its steps requires prices or a forecasting model, and all of them can be carried out before any model is fitted.

\begin{enumerate}
\item \textbf{Store publication and finalisation times for every run.} The issue time identifies a run but says nothing about when it could be used. If a vendor does not supply publication and finalisation times, record them by polling. Our live check, limited to three runs on one evening, supports this at one-minute resolution.
\item \textbf{Define gates in the market's local clock time.} Convert each gate to UTC for each delivery day, so that daylight-saving transitions are reflected.
\item \textbf{Choose an explicit action margin $\Delta$.} State it, and report how the selected runs change across a plausible range of margins (Table 5). The relevant thresholds can be as short as 5 minutes.
\item \textbf{Select runs by the finalisation rule,} as a conservative and auditable choice. Use the publication rule as a sensitivity check. Where both rules select the same runs, results cannot depend on values that changed after the gate; where they select different runs, report how much the results differ.
\item \textbf{Recompute delays periodically and monitor input freshness.} Publication delays changed by up to 1.4 hours within our sample, so a rule calibrated on past delays can silently start to leak. Track the age of the selected runs over time, and retrain or report when it shifts.
\item \textbf{Treat outturn series separately.} Either lag them conservatively (here, revisions were observed up to 32 days after delivery, a lower bound), or retain the first-published values from the start of data collection. Report which choice was made and the revision exposure that remains.
\item \textbf{Record forecasts at each decision point from now on.} Shortly before each gate, store the latest published runs together with their values. After some months this yields the as-of-gate values that the archive cannot provide.
\item \textbf{Report all of the above in the data section,} together with the share of delivery days on which the naive rule would have selected a different run. This number can be computed from timestamps alone and gives readers a direct measure of the exposure. Where values are available, also report how much the selected inputs differ (Section 5.5).
\end{enumerate}

With access to an archive that records publication and finalisation times, the same audit can be repeated for other vendors, markets and periods.

\section{Limitations and future work}

The empirical results concern one vendor's archive, three forecast variables, two auction gates in one market, and the period 2022--2026. The mechanisms we describe are general: the difference between issue and publication time, the treatment of local clock time, drifting publication delays and unrecorded revisions. Their magnitudes, however, are specific to this setting. Another vendor, market or period will show different leakage rates. This is the main reason we recommend auditing each dataset rather than relying on the figures reported here.

The vendor's documentation of the three timestamps is minimal. Our interpretation rests on the regularity of the observed delays and on a direct check over one evening, which confirmed it for three runs. A longer live record would strengthen it.

The publication rule identifies runs that were available at the gate, but not the values they had at the gate, since the archive stores only finalised values. The finalisation rule avoids this problem at the cost of using older runs. Because finalisation is recorded per run rather than per forecast step, it may also reject runs whose delivery-day values were already settled at the gate, so it is likely stricter than necessary. We have not recovered the as-of-gate values themselves.

Finally, this paper measures the exposure of input selection and its effect on the inputs themselves, not its consequences for trading. Section 5.5 shows that the wrongly selected inputs differ materially from those available at the gate and that their apparent accuracy can be higher or lower, but it compares archived values against revised outturns, for three deterministic configurations over 2024--2026. The cause of the daily accuracy cycle in the vendor's demand forecasts is also not established. The natural next step is to measure the \emph{economic} cost: how much a trading strategy's apparent profitability changes when the naive rule is replaced by a point-in-time rule, and what each minute of margin costs. We leave this to future work.

\section{Conclusion}

Point-in-time correctness is easily treated as a property that a dataset either has or lacks, checked once. Our results suggest that it is better understood as a dynamic property of the data pipeline. Whether a selection rule leaks depends on the weather-model configuration, the gate and the season, and it can switch abruptly at daylight-saving transitions, when a vendor's processing changes, and when a participant's action margin crosses a threshold of a few minutes. The effect on the inputs is material but of no predictable sign. Three pitfalls produce these effects, and drifting delays make every static fix fragile. The remedy is procedural rather than statistical: audit the past from timestamps, record the future as it arrives, and report both.

\enlargethispage{4\baselineskip}\section*{References}

{\footnotesize\setlength{\parskip}{0pt}\begin{list}{}{\leftmargin=1.2em\itemindent=-1.2em\itemsep=1pt\parsep=0pt}
\item Croushore, D. (2011). Frontiers of real-time data analysis. \emph{Journal of Economic Literature}, 49(1), 72--100. \href{https://doi.org/10.1257/jel.49.1.72}{doi:10.1257/jel.49.1.72}
\item Croushore, D., \& Stark, T. (2001). A real-time data set for macroeconomists. \emph{Journal of Econometrics}, 105(1), 111--130. \href{https://doi.org/10.1016/S0304-4076(01)00072-0}{doi:10.1016/S0304-4076(01)00072-0}
\item ECMWF (2026). \emph{Dissemination schedule}. ECMWF Confluence, Data and Charts documentation, last updated 7 October 2026. Retrieved 14 October 2026 from {\footnotesize\url{https://confluence.ecmwf.int/spaces/DAC/pages/272310483/Dissemination+schedule}}
\item EPEX SPOT (2024). \emph{Trading at EPEX SPOT 2024} (trading brochure), July 2024. Retrieved 14 October 2026 from {\footnotesize\url{https://www.epexspot.com/sites/default/files/2024-07/Trading%20brochure%20July%202024.pdf}}
\item Kapoor, S., \& Narayanan, A. (2023). Leakage and the reproducibility crisis in machine-learning-based science. \emph{Patterns}, 4(9), 100804. \href{https://doi.org/10.1016/j.patter.2023.100804}{doi:10.1016/j.patter.2023.100804}
\item Kaufman, S., Rosset, S., Perlich, C., \& Stitelman, O. (2012). Leakage in data mining: Formulation, detection, and avoidance. \emph{ACM Transactions on Knowledge Discovery from Data}, 6(4), Article 15, 1--21. \href{https://doi.org/10.1145/2382577.2382579}{doi:10.1145/2382577.2382579}
\item Künsch, H. R. (1989). The jackknife and the bootstrap for general stationary observations. \emph{The Annals of Statistics}, 17(3), 1217--1241. \href{https://doi.org/10.1214/aos/1176347265}{doi:10.1214/aos/1176347265}
\item Lago, J., Marcjasz, G., De Schutter, B., \& Weron, R. (2021). Forecasting day-ahead electricity prices: A review of state-of-the-art algorithms, best practices and an open-access benchmark. \emph{Applied Energy}, 293, 116983. \href{https://doi.org/10.1016/j.apenergy.2021.116983}{doi:10.1016/j.apenergy.2021.116983}
\item Narajewski, M., \& Ziel, F. (2020). Econometric modelling and forecasting of intraday electricity prices. \emph{Journal of Commodity Markets}, 19, 100107. \href{https://doi.org/10.1016/j.jcomm.2019.100107}{doi:10.1016/j.jcomm.2019.100107}
\item National Grid NSL (2021). \emph{NSL access rules and charging methodology}, version 1.0, June 2021. Retrieved 14 October 2026 from {\footnotesize\url{https://www.ofgem.gov.uk/sites/default/files/2021-08/NSL%20Access%20Rules%20and%20Charging%20Methodology.pdf}}
\item Nord Pool (2020). \emph{Nord Pool announces GB auction changes for Brexit}. Exchange message, 19 October 2020. Retrieved 14 October 2026 from {\footnotesize\url{https://www.nordpoolgroup.com/en/message-center-container/newsroom/exchange-message-list/2020/q4/nord-pool-announces-gb-auction-changes-for-brexit/}}
\item Nord Pool (2023). \emph{Product specifications (GB market)}, effective 10 July 2023. Retrieved 14 October 2026 from {\footnotesize\url{https://www.nordpoolgroup.com/4a3355/globalassets/download-center/rules-and-regulations/product-specifications-gb-valid-from-10-july-23.pdf}}
\item Politis, D. N., \& Romano, J. P. (1992). A circular block-resampling procedure for stationary data. In R. LePage \& L. Billard (Eds.), \emph{Exploring the Limits of Bootstrap} (pp. 263--270). Wiley, New York.
\item Reinert, D., Prill, F., Frank, H., Denhard, M., Baldauf, M., Schraff, C., Gebhardt, C., Marsigli, C., Förstner, J., Zängl, G., Schlemmer, L., Blahak, U., \& Welzbacher, C. (2026). \emph{DWD Database Reference for the Global and Regional ICON and ICON-EPS Forecasting System}, version 2.5.6. Deutscher Wetterdienst, Offenbach am Main. Retrieved 14 October 2026 from {\footnotesize\url{https://www.dwd.de/DWD/forschung/nwv/fepub/icon_database_main.pdf}}
\item Sheng, Z., Yuan, Y., Zhang, G., \& Li, Y. (2026). R$^2$Energy: A large-scale benchmark for robust renewable energy forecasting under diverse and extreme conditions. arXiv:2602.15961.
\item Snodgrass, R. T. (1999). \emph{Developing Time-Oriented Database Applications in SQL}. Morgan Kaufmann, San Francisco. ISBN 1-55860-436-7.
\item Stark, T., \& Croushore, D. (2002). Forecasting with a real-time data set for macroeconomists. \emph{Journal of Macroeconomics}, 24(4), 507--531. \href{https://doi.org/10.1016/S0164-0704(02)00062-9}{doi:10.1016/S0164-0704(02)00062-9}
\item Weron, R. (2014). Electricity price forecasting: A review of the state-of-the-art with a look into the future. \emph{International Journal of Forecasting}, 30(4), 1030--1081. \href{https://doi.org/10.1016/j.ijforecast.2014.08.008}{doi:10.1016/j.ijforecast.2014.08.008}
\end{list}}
\end{document}